\documentclass[12pt,a4paper]{article}

\usepackage[british]{babel}

\usepackage[a4paper,top=2cm,bottom=2cm,left=2.5cm,right=2.5cm,marginparwidth=1.75cm]{geometry}

\usepackage[style=apa, backend=biber]{biblatex} 
\DeclareLanguageMapping{british}{british-apa} 
\DeclareFieldFormat[article]{volume}{\apanum{#1}} 

\usepackage{amsmath}
\usepackage{graphicx}
\usepackage[colorlinks=true, allcolors=blue]{hyperref}
\usepackage{hyperref}
\usepackage[title]{appendix}
\usepackage{mathrsfs}
\usepackage{amsfonts}
\usepackage{booktabs} 
\usepackage{caption}  
\usepackage{threeparttable} 
\usepackage{algorithm}
\usepackage{algorithmicx}
\usepackage{algpseudocode}
\usepackage{listings}
\usepackage{enumitem}
\usepackage{chngcntr}
\usepackage{booktabs}
\usepackage{lipsum}
\usepackage{subcaption}
\usepackage{authblk}
\usepackage[T1]{fontenc}    
\usepackage{csquotes}       
\usepackage{diagbox}

\usepackage{setspace}
\usepackage{titlesec}
\titleformat{\section} 
  {\normalfont\Large\bfseries}{\thesection.}{1em}{}
  
\usepackage{float}   
\usepackage{caption} 


\title{MathLearner: A Large Language Model Agent Framework for Learning to Solve Mathematical Problems}

\author[1]{Wenbei Xie}
\author[1, *]{Donglin Liu}
\author[1, *]{Haoran Yan}
\author[1]{Wenjie Wu}
\author[1]{Zongyang Liu}
\affil[1]{\small Beijing-Dublin International College, Beijing University of Technology, 100 Pingleyuan, Chaoyang District, 100124, Beijing, China}
\affil[*]{\small These authors contributed equally to this work.}

\begin{document}
\maketitle

\begin{abstract}
With the development of artificial intelligence (AI), large language models (LLM) are widely used in many fields. However, the reasoning ability of LLM is still very limited when it comes to mathematical reasoning. Mathematics plays an important role in all aspects of human society and is a technical guarantee in the fields of healthcare, transport and aerospace, for this reason, the development of AI big language models in the field of mathematics has great potential significance. To improve the mathematical reasoning ability of large language models, we proposed an agent framework for learning to solve mathematical problems based on inductive reasoning. By emulating the human learning process of generalization of learned information and effective application of previous knowledge in new reasoning tasks, this framework has great performance in the mathematical reasoning process. It improves global accuracy over the baseline method (chain-of-thought) by 20.96 \% and solves 17.54 \% of the mathematical problems that the baseline cannot solve. Benefiting from the efficient RETRIEVAL method, our model improves the ability of large language models to efficiently use external knowledge, i.e., the mathematical computation of the model can be based on written procedures. In education, our model can be used as a personalised learning aid, thus reducing the inequality of educational resources.
\end{abstract}



\section{INTRODUCTION}

With the integration of artificial intelligence (AI) and natural language processing (NLP) technologies into people's daily lives, the demand for more advanced and capable language models has become imperative. These language models, often referred to as Large Language Models (LLMs) have demonstrated remarkable capabilities in dealing with automating intricate tasks, such as understanding and generating human-like text \cite{OpenAI_2023}. In the real world, LLMs have been applied in various applications, including AI chatting systems like ChatGPT to intelligent coding assistant tools like GitHub Copilot \cite{OpenAI_2023, github_copilot}.

However, one area where their potential remains largely untapped is in the domain of mathematics. Mathematics holds immense significance in various aspects of daily life, serving as the foundation for fields such as science, engineering, finance, and technology. Mathematics holds immense significance in various aspects of daily life, serving as the foundation for fields such as science, engineering, finance, and technology. It provides the language and tools necessary for understanding and modelling natural phenomena, designing innovative technologies, and making informed decisions in business and finance. From calculating trajectories in space exploration to optimizing algorithms in computer science, mathematics permeates every aspect of modern society. However, despite the importance of mathematics, language models have not been extensively explored in this domain. Unlike other types of problems, mathematical tasks demand precise and complex reasoning, pattern recognition, and algorithmic thinking, posing unique challenges for language models \cite{math_importance}. The ability to accurately solve mathematical problems is crucial not only for academic success but also for practical applications in fields such as engineering, finance, and data analysis. Therefore, there is an impending demand to enhance the reasoning ability of LLMs through training and testing using Math problems to fit in the growing needs for applications.

Previous efforts to further improve the reasoning ability of LLMs, such as Chain of Thought (CoT) and Retrieval-Augmented Generation (RAG), have made a significant impact in enhancing the problem-solving capabilities \cite{NEURIPS2022_9d560961, lewis2020retrieval}. CoT, for instance, focuses on guiding language models using a series of reasoning steps to arrive at more accurate solutions, while RAG leverages retrieval-based techniques to augment generation, enabling LLMs to learn solutions from existing data.
Although these approaches have made promising results, they also exhibit limitations in solving problems that are similar to the problems seen before. For example, previous RAG studies have shown that solutions discovered by LLMs may suffer from overfitting, particularly when relying on traditional retrievers like BM-25 or Density Paragraph Retrievers (DPRs) \cite{prad2023address}. These traditional retrievers normally rely on keyword match, which will lead to solutions that have exactly the same keywords as the query problem and struggle to generalize to similar but unseen problems. Consequently, there is a need for more advanced techniques that can address the limitations of existing approaches and provide a generalizing ability to solve similar questions across diverse problem domains.

To address the limitations of previous approaches, we have developed a novel framework called MathLearner, inspired by the principles of human learning, particularly inductive reasoning. Human learning often involves inferring general principles or solutions from specific examples and applying this knowledge to novel situations. Similarly, MathLearner aims to empower LLMs to learn to resolve math problems by leveraging inductive reasoning principles.

MathLearner operates in three main stages, mirroring the stages of human learning:
\begin{enumerate}
    \item Learning from Examples: The framework begins by exposing the LLM to a diverse set of math problems and their solutions, allowing it to learn from annotated examples.
    \item Memorizing Solving Methods: MathLearner then focuses on memorizing various problem-solving methods and techniques, enabling the LLM to build a repository of strategies for tackling different types of math problems. approach problem-solving in a more systematic and structured manner.
    \item Recalling Previous Knowledge: Finally, MathLearner equips the LLM with the ability to recall and apply its learned knowledge to solve new math problems, mimicking the process of retrieving and applying previously learned solutions.
\end{enumerate} 

Through these principles and design, MathLearner aims to not only enhance the reasoning ability of LLMs but also to enable continuous and instant learning, ultimately improving their ability to solve math problems accurately and efficiently. The potential impact of enhancing LLMs’ math problem-solving abilities extends beyond the realm of AI research to education and real-world applications. By improving LLMs’ ability to solve math problems, MathLearner can lower the barrier to self-study among students, providing a more diverse learning platform, and facilitating easier review of incorrect answers. This can revolutionize education by empowering students to learn independently, offering varied learning experiences, and simplifying the process of reviewing mistakes.

The dataset we used to train and test is MATH \cite{hendrycks2021measuring}. It has many challenging mathematics problems that are difficult for LLMs to solve in different categories, such as algebra and geometry, which is useful for us to train and test the problem-solving skills of LLMs for different categories of problems. In addition to that, one of the key strengths of the MATH dataset is its provision of step-by-step solutions for every problem, offering detailed insights into the reasoning processes required to arrive at the correct solutions. By leveraging the step-by-step solutions provided in the dataset, we can effectively train the model to emulate human-like problem-solving approaches and improve its overall performance.

In summary, this paper presents several key contributions:
\begin{itemize}
    \item We propose a new retrieval method based on features to retrieve solutions to similar problems for the encountered problem.
    \item We design a learning framework which can effectively reuse previously learned knowledge to solve current problems.
\end{itemize}

\section{RELATED WORK}
In the field of enhancing LLM reasoning ability, numerous studies have been conducted to explore the area of LLM reasoning. Step-by-step reasoning, think-in-memory, and few-shot learning are three hot topics to address this issue. Our framework implements these three concepts to varying degrees, improving LLM's mathematical reasoning. The following is an introduction to these three concepts and related research.

\textbf{Step-by-step reasoning} Many works indicate that step-by-step reasoning improves LLM performance \cite{k_level, kg_agent, cumulative, learn_from_explanation, semi-structured, wen2024mindmap}and correspondingly, that this performance can be further enhanced with appropriate guidance and use of tools \cite{cumulative, semi-structured, wen2024mindmap}. Mingyu and his team found that chain of thought could significantly improve the predictive accuracy of the LLM's reasoning across multiple datasets \cite{jin2024impact}. They stated that longer reasoning chains containing misleading information still enhance the reasoning performance of LLMs, indicating that chain length is more vital than accuracy for problem-solving. Zhang et al. and Zelikman et al. also point out that LLMs with step reasoning progress can thoroughly and effectively utilise existing conditions to derive novel understandings \cite{k_level, zelikman2023parsel}. Wei et al. first introduced the concept of chain of thought (CoT), a series of intermediate reasoning steps in 2022\cite{NEURIPS2022_9d560961}. Experiments showed that normalised thought chain cues greatly improved the ability of large language models to perform complex reasoning across a range of arithmetic, commonsense and symbolic reasoning tasks. Later Lyu et al. proposed the faithful CoT (FCoT) based on CoT. Unlike CoT, faithful CoT reflects how the model arrives at the answer, which improves the empirical performance of LLM\cite{lyu2023faithful}. It outperforms standard CoT on 9 of 10 benchmarks from 4 diverse domains, with a relative accuracy gain of 6.3\% on Math Word Problems (MWP), 3.4\% on Planning, 5.5\% on Multi-hop Question Answering (QA), and 21.4\% on Relational Inference. These findings and achievements inspire us to enhance the mathematical problem-solving capabilities of Large Language Models (LLMs) through step-by-step learning and reasoning applications.

\textbf{Think-in-Memory} Thinking in memory addresses LLM repetitive reasoning using locality-sensitive hashing in dialogue to retrieve long-term memory efficiently. \cite{thinking_memory, kg_agent, luo2024reasoning}Referring to useful historical information not only reduces the process of repetitive reasoning but also increases the accuracy of the results. Knowledge graphs (KG) and vector databases (VecBD) are two methods that are widely used to store helpful information\cite{}. KG-based LLM reasoning uses embedding space to represent the entities and special model architectures for the reasoning process \cite{kg_agent, reasoning_on_graph, reasoning_with_kg_and_llm, zhang2022greaselm}. One earlier work is by Logan IV et al., where they proposed mechanisms for a neural language model to select and incorporate facts from a knowledge graph relevant to the text context, enhancing the factual generation capabilities of LLMs \cite{logan2019baracks}. In the meantime, it is necessary to determine an effective method for fusing and reasoning over the KG representations and the language context, which provides situational constraints and nuances. Further research has led to the development of techniques to enhance the model's capability for utilizing the knowledge graph (KG). Abu-Rasheed et al. put forth an approach for LLM prompts to reduce the risk of model hallucinations and safeguard against including erroneous or imprecise information in 2024 \cite{aburasheed2024knowledge}. In 2024, Jiang et al. proposed KG-Agent\cite{kg_agent}, an autonomous LLM-based agent framework. This enables a small LLM to actively make decisions over the reasoning process aid with KGs. 
However, in terms of the speed of access to information, VecDBs have the advantage over KG \cite{mittal2017thinking}, and VecDBs are well-capable for retrieval applications \cite{asai-etal-2023-retrieval}. VecBD-based LLM reasoning uses vector databases to build specialized embeddings for special category information with domain-specific embeddings to support efficient information retrieval during the processing progress \cite{vector}. Numerous studies and practical applications have shown the significant impact of VecDBs. For instance, Azure AI Search, previously known as Cognitive Search, enhanced its vector search capabilities using Qdrant\cite{qdrant2023}. Similarly, Pinecone's Canopy\cite{pineconeCanopy}, an open-source framework, utilizes Pinecone’s VecDB to develop and deploy ready-to-use RAG systems. Given that vector databases excel in efficiently retrieving similar information, our framework leverages vector databases for storing learning histories. This think-in-memory method facilitates more efficient information utilization in the User module of our framework, optimizing our system's performance by harnessing the strengths of vector databases for rapid and relevant data access during the think-in-memory process.

\textbf{Few-shot learning} It is a method that can predict new classes when only one or a few labels are available for each class of training data \cite{one_shot, few_shot}. This method is particularly valuable in scenarios where data collection costs are high, data is scarce, or when there is a need for models to adapt to new tasks quickly. \cite{one_shot}. Powerful few-shot learning capability also leveraged LLMs to perform reasoning over KG.  In our research, we implement this method with GPT-4.0 to address the high costs in time and resources required for processing queries, typically 8-10 tokens and 1-2 minutes each. This method allows us to minimize expenses while achieving comparable results to extensively annotated methods. By carefully selecting and crafting prompt examples, we guide the model to generalize from minimal input efficiently, leveraging GPT-4.0’s advanced training. This approach not only reduces costs but also maintains high-quality outcomes, demonstrating the effectiveness of few-shot learning in resource-constrained scenarios.

By applying these three concepts to our framework, the mathematical reasoning of our framework is enhanced. more details in the framework will be elaborated in the methodology section.

\section{METHODOLOGY}
In this section, we provide an overview of the MathLearner framework. First, we describe the problem that this paper desires to solve more accurately. After that, a detailed description of the learning module is presented. Finally, we introduce the structure of the application module. Specifically, a further discussion is made about the generation of sets of features to describe questions, which is a general challenge in the framework.
\subsection{Problem Statement}
We consider a basic pattern of human learning: For a category of mathematical problems, one or several problems and their solutions are first given to a student. Based on these examples the students generalize a general solution to the type of mathematical problems. Besides, the student would remember a set of generalized features of the given problems. When the student encounters a new problem, the student would also use a set of previously learned features to describe the problem. By searching through the memory using feature match, the student may find a useful solution that previously learned to solve the new problem. Thus, our objective is to use LLMs to simulate the whole procedure of human learning and solve mathematical problems. We noticed that the whole pattern mentioned above can be divided into two parts, learning and applying. Therefore, the MathLearner framework contains two modules to simulate the pattern.

\begin{figure}[h]
    \centering
    \includegraphics[width=1\textwidth]{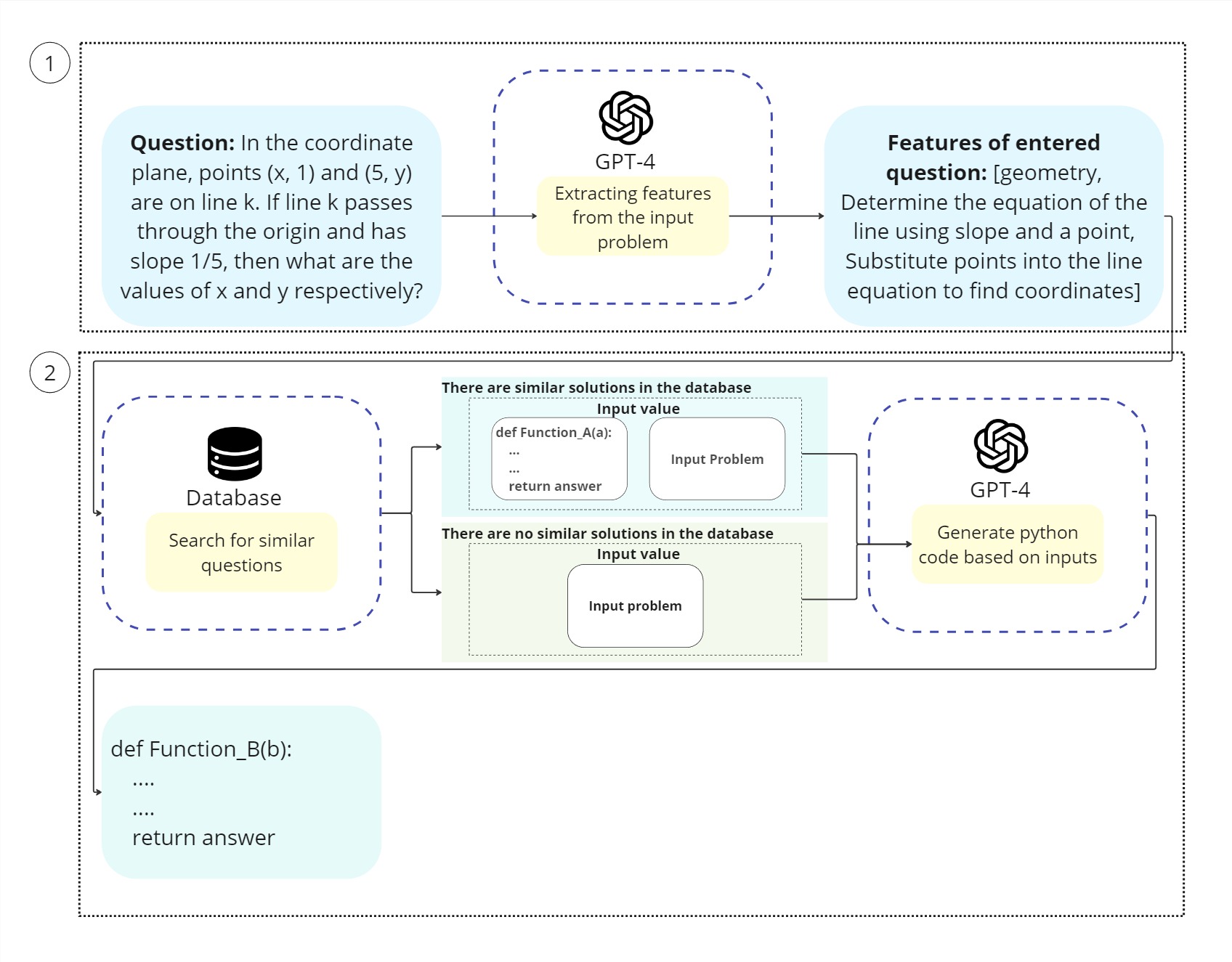}
    \caption{a case study on using MathLearner, which performs two main steps: extracting feature values from the input problem and finding similar solution steps in the database based on these feature values. If similar solution steps exist in the database, we send these steps along with the new input problem to GPT so that GPT can generate Python code to solve the problem. If similar problem-solving steps do not exist in the database, we send the problem directly to GPT, allowing it to generate the Python code that solves it. This process makes the problem-solving process more efficient and precise.}
    \label{fig:enter-label}
\end{figure}

\subsection{Pipeline of Learning Module}
\begin{figure}[h]
    \centering
  \includegraphics[width=0.8\linewidth]{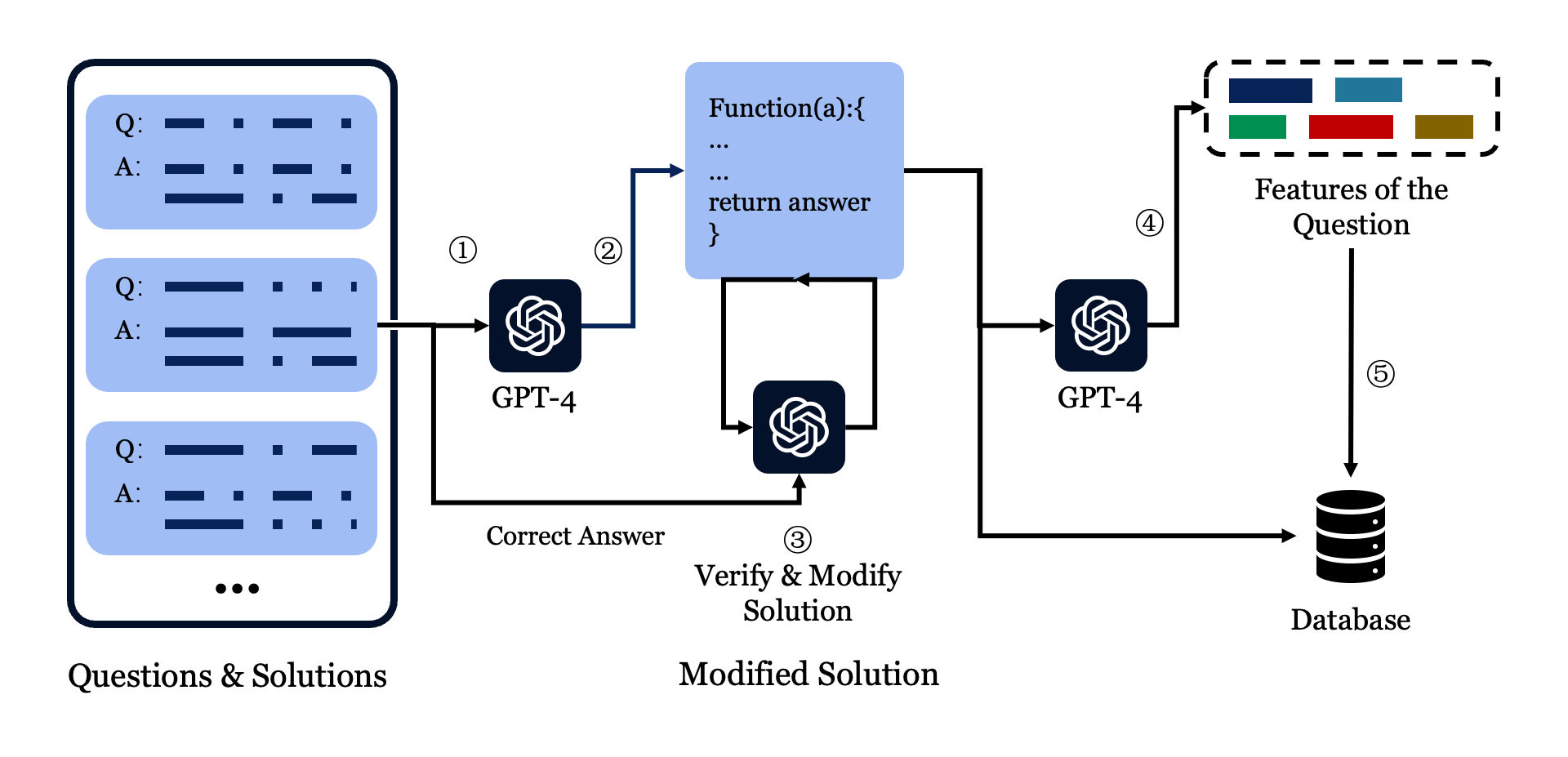}
  \caption{An overview of the pipeline of Learning Module. From left to right, the large language model 1) receives examples, 2) generates modified solutions, 3) verifies and modifies possible errors in the solution, and 4) generates features for questions. Then, the features and solution will be saved into a database.}
  \label{fig:learning_module}\
\end{figure}
\subsubsection{Modified Solutions Generation and Verification}
As demonstrated in Figure 3.1, the second step of the pipeline is to generate modified solutions. To effectively be useful for future problem solving, the modified solutions should be in the form of a program. Therefore, we define a sub-pipeline for generating the modified solutions under the instruction of Parsel, an efficient program generation method. First, LLM is asked to divide the given solution into steps. Then, several Parcel programs, which can be understood as a pseudo-code, are generated based on the steps in the divided solution. Finally, the Parsel programs are translated into one single real program containing multiple functions. Compared to the original Parsel implementation, we decided to enter solutions with questions to simulate the human learning process better. As the third step, we also learn the validation method from Parsel. After one modified solution is generated, the program in the solution would be run to check the correctness of the solution. If it does not pass, the LLM will be required for a new version of a modified solution.
\subsubsection{Feature Generation and Storage}
Consider two mathematical problems with similar solutions, one problem is a word problem, which means the information on the problem is presented in ordinary language, while another problem is expressed by mathematical notation. To match these two problems into one category can be a challenge by using traditional retrieval methods, which are generally based on keyword match. Thus, a novel retrieval method should be developed to match problems that have similar solutions.
As mentioned before, humans will store the feature representation of the learned problems. In detail, humans will remember some features for each step of the solution for the problem. When they encounter a similar problem, a set of features will also generated for each of the possible steps of the new solution. These features will be used to retrieve previously learned solutions. To simulate this procedure, the Learning module will ask the LLM to generate two types of features. One is a general description of the type of problem, like algebra or geometry. The other features are used to describe the operations or theorems used in each of the steps. After this, the set of features will be translated into vectors and stored in a vector database for the similarity search in the future.
\subsection{Pipeline of Application Module}
\begin{figure}[h]
    \centering
  \includegraphics[width=0.8\linewidth]{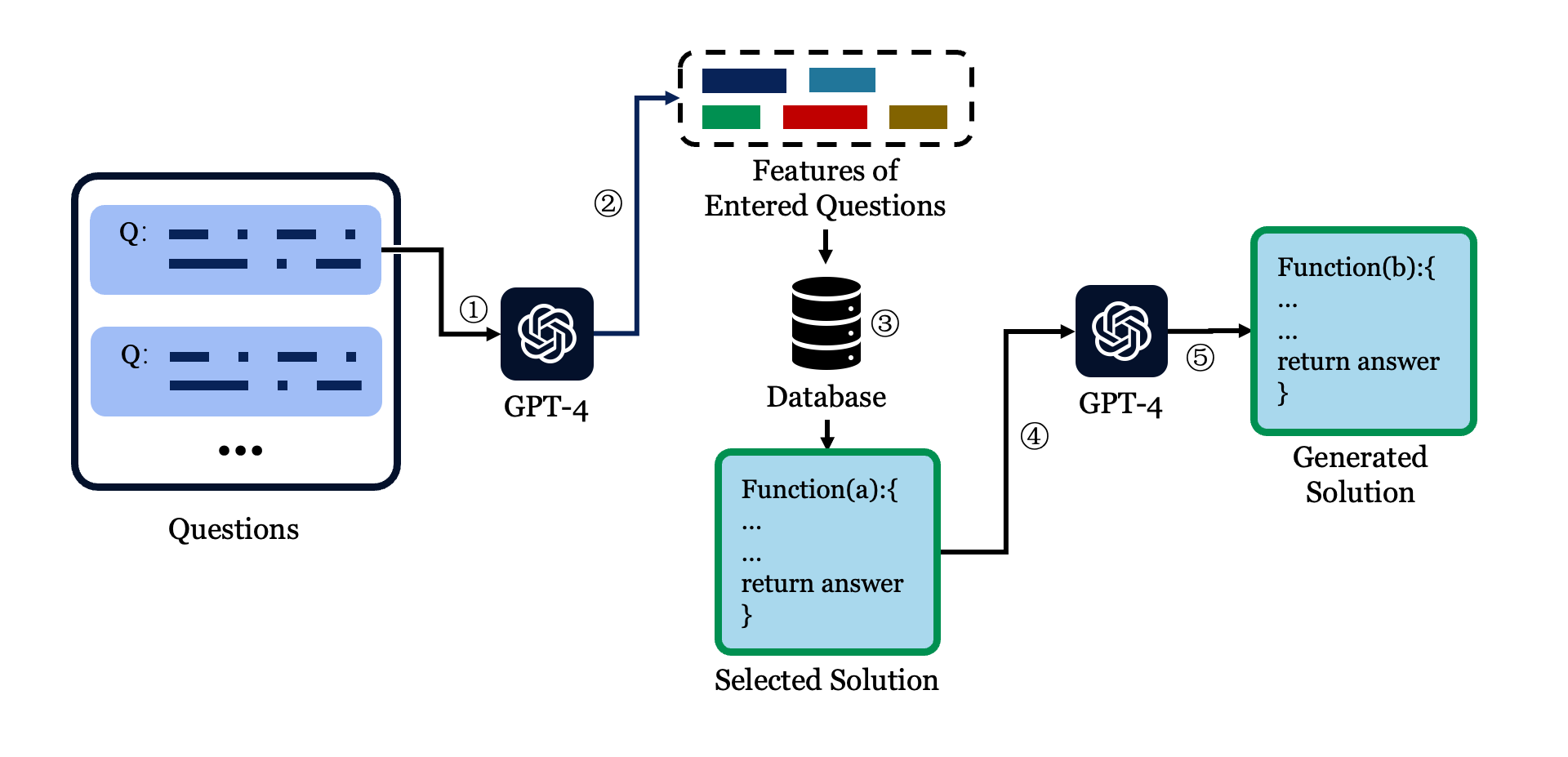}
  \caption{An overview of the pipeline of Application Module. From left to right, the large language model 1) receives questions, 2) generates features for entered questions, 3) retrieves the closest solution, and 4) generates new solutions for entered questions.}
  \label{fig:learning_module}'\
\end{figure}

\subsubsection{Extracting Features from Problems}
As demonstrated in Figure 3.2, the second step of the pipeline involves extracting two types of features from the provided problem: a general description of the problem and the operations or theorems involved in each step. In step three, these extracted features are converted to vector form and used to perform a vector search of the database. The purpose of this step is to quickly match related problems in the database and find similar solutions. This approach allows LLM to find similar problems faster and to write the code to solve the problem from the code stored in the database. This process mimics human problem-solving behaviour by identifying a characteristic description of the problem and solving it using the appropriate operations or theorems.

\subsubsection{Feature matching and answer generation}
Matching of features can be done through vector search in the third step. When the feature matching is successful, the solution ideas stored in the database are sent to the LLM along with the topic so that it draws on the stored solution ideas to generate solution ideas for the new problem. This process, which draws on the solution ideas in the database, helps to improve the correctness of the newly generated solution ideas and speeds up the generation of solution ideas by the LLM. In this step, we took into account the fact that when the problem contains completely new features, it means that this step will not match what is in the database. Therefore, we take the same approach as the training sub-pipeline and let the LLM generate the corresponding solution. This maximizes the correctness of the solutions.

\subsection{Evaluation Metrics}
The effectiveness of the MathLearner framework was quantitatively assessed using a suite of rigorously defined metrics. For comparison, the Chain of Thought (CoT) framework served as the baseline. The CoT approach involves breaking down complex problems into simpler, sequential steps, thereby facilitating deeper reasoning and understanding. This method has proven effective in enhancing the problem-solving capabilities of language models by guiding them through a structured thought process.

The evaluation metric we used is also from MATH dataset. We randomly select 150 questions from precalculus section, which is useful to test the performance of our system.

In our study, we categorized each test question into one of four quadrants based on whether a similar problem was retrieved and whether the final calculated result was correct (see Figure 3.3). These quadrants serve as a framework for evaluating the performance of MathLearner and are instrumental in calculating the various metrics detailed in the subsequent sections.

\begin{figure}[h]
    \centering
    \includegraphics[width=0.3\textwidth]{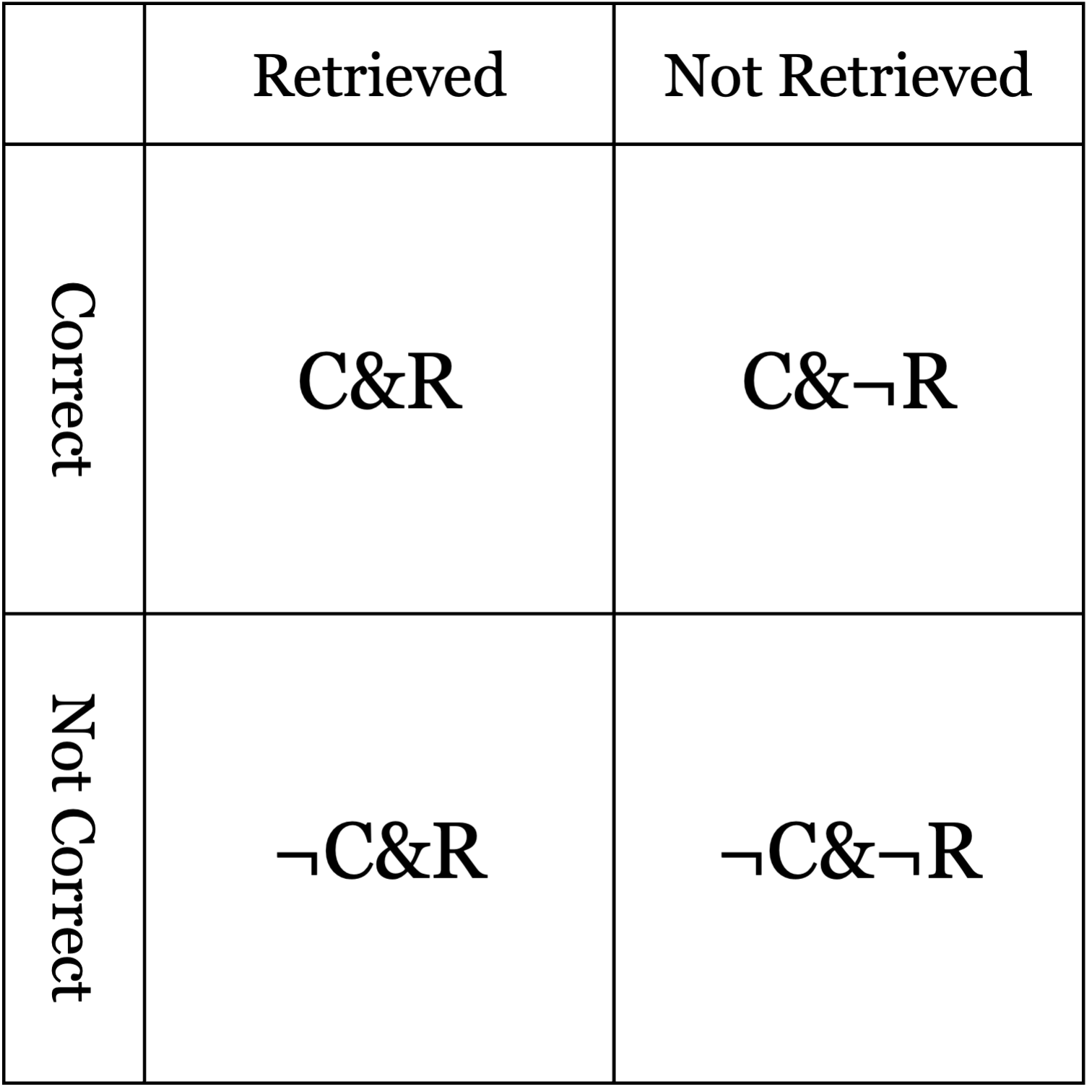}
    \caption{The category of the questions ($U = C\&R + C\&\neg R + \neg C\&R + \neg C\& \neg R$)}
    \label{fig:enter-label}
\end{figure}

\begin{itemize}
    \item Global Accuracy: Global Accuracy measures the proportion of correctly solved problems among all attempted problems. MathLearner's overall accuracy is calculated as the ratio of correctly solved problems to the total number of problems. This metric reflects the proportion of problems correctly solved across all attempts and serves as a primary indicator of the system's overall effectiveness. This metric is scaled between 0 and 1, where 0 indicates no problems solved correctly, and 1 represents perfect performance across all problems.
    \begin{equation}
    \text{Global Accuracy} = \frac{\text{Number of Correct Solutions ($C\&R + C\&\neg R$)}}{\text{Number of All Question in the Dataset ($U$)}} , [0, 1]
    \end{equation}
    \item Accuracy Contribution: This metric evaluates the effectiveness of MathLearner in finding similar results and producing correct solutions. Specifically, it measures the proportion of problems where MathLearner finds similar results and provides correct solutions among all correctly solved problems. It is calculated by:
    \begin{equation}
        \text{Accuracy Contribution} = \frac{\text{Correct Solutions using Similar Solution ($C\&R$)}}{\text{Number of Correct Solutions ($C\&R + C\&\neg R$)}} , [0, 1]
        \end{equation}
    \begin{itemize}
        \item Profitability (Benefit): Profitability quantifies the extent to which finding similar results contributes to correct solutions. It is calculated as the ratio of global accuracy to the accuracy achieved by CoT, minus 1. It is calculated by:
         \begin{equation}
        \text{Profitability} = \frac{\text{Global Accuracy}}{\text{CoT Global Accuracy}}-1 , [0, +\infty)
        \end{equation}
        A higher profitability indicates a higher times contribution to finding similar results to correct solutions than the baseline.

    \end{itemize}
    \item Precision Accuracy: This metric assesses the accuracy of the system in scenarios where the questions find a similar solution. This metric is crucial for applications that require high precision and is calculated as:
        \begin{equation}
        \text{Precision Accuracy} = \frac{\text{Correct Solutions using Similar Solution ($C\&R$)}}{\text{Number of Problem with Similar Solution ($C\&R + \neg C\& R$)}} , [0, 1)
        \end{equation}
    A higher score on this metric reflects the system's ability to handle precise queries effectively. When the training set is large enough, most of the problems should find a similar solution, making the global accuracy approach the precision accuracy.
    \item Target Achievement Rate: The definition of the target is the framework is expected to help LLM to answer all the problems using retrieved solutions, which requires the framework can correctly answer all the problems which CoT failed to do. Thus, the goal achievement rate can be calculated using the following formula:

    \begin{equation}
        \text{Target Achievement Rate} = \frac{\text{MathLearner Correct Solutions} - \text{CoT Correct Solutions}}{\text{Total Number of CoT Unresolved Problems}} , [0, 1]
    \end{equation}
     Here, a score of 1 signifies that the framework successfully resolved every problem that CoT failed to solve, while a score of 0 would mean it failed to solve any additional problems.

\end{itemize}

By employing these evaluation metrics, we aim to provide a comprehensive assessment of MathLearner's performance, considering not only its accuracy but also its effectiveness in finding similar results and achieving correct solutions across various problem domains.

\section{Results}

We choose "Chain-of-Thought" as our baseline. In this baseline, we do not use any prompts and let GPT generate the Python code to solve the problem in a step-by-step approach. In this case, the probability that the code output by GPT can solve the problem is defined as CoT.

\begin{table}[!ht]
    \centering
    \caption{Performance of Chain-of-Thought (CoT)}
    \begin{tabular}{|>{\centering\arraybackslash}m{2cm}|>{\centering\arraybackslash}m{2cm}|>{\centering\arraybackslash}p{10cm}|}
    \hline
        Evaluation Metrics & Value & Description \\ \hline
        Global Accuracy & 41.33\% & $\frac{\text{Number of Precisely Correct Solutions using CoT}}{\text{Number of All Quetion in the Dataset}} = 0.4133 $ \\ \hline
    \end{tabular}
\end{table}

 We obtained the relevant test results for our program through multiple tests. We have used four evaluation criteria to show the performance and accuracy of our program. As shown in the table below:

\begin{table}[!ht]
    \centering
    \caption{Performance of MathLearner}
    \begin{tabular}{|>{\centering\arraybackslash}m{2cm}|>{\centering\arraybackslash}m{2cm}|>{\centering\arraybackslash}p{10cm}|}
    \hline
        Evaluation Metrics & Value & Description  \\ \hline
        Global Accuracy & 50\% & $\frac{\text{Number of Correct Solutions using MathLearner}}{\text{Number of All Question in the Dataset}} = 0.5$ \\ \hline
        Profitability (Benefit) & 20.96\% & $\frac{\text{Global Accuracy}}{\text{CoT Global Accuracy}}-1 = 0.2096$  \\ \hline
        Precision Accuracy & 51.55\% &  $\frac{\text{Correct Solutions using Similar Solution}}{\text{Number of Problem with Similar Solution}} = 0.5155$ \\ \hline
        Target Achievement Rate & 17.54\% & $\frac{\text{MathLearner Correct Solutions} - \text{CoT Correct Solutions}}{\text{Total Number of CoT Unresolved Problems}} = 0.1754$ \\ \hline
    \end{tabular}
\end{table}

\begin{itemize}
    \item Global Accuracy (Overall): By using our MathLearner for problem-solving, we successfully
    solved 75 out of 150 problems. This result demonstrates the overall performance of MathLearner in problem-solving, which includes solving unknown problems, i.e., problems that have not been studied, as well as problems that have been studied similarly.
    \item Profitability (Benefit): This metric quantifies the extent to which the program contributes to having learned a solution to a similar problem, reflecting how effectively our program uses existing knowledge to find the right solution.
    \item Precision Accuracy: This metric indicates how effectively the system finds and solves problems when similar results are available, it can be considered an important indicator of the correctness of our program.
    \item Target Achievement Rate: Measures the framework's effectiveness in enabling LLM to use retrieved solutions to correctly answer all the problems which CoT failed to address.
\end{itemize}

By comparing the global accuracy of MathLearner (50\%) with the global accuracy of CoT (41.33\%), we can observe a significant improvement in accuracy with the use of MathLearner, approximately an increase of 10\%. In addition, we believe that precision accuracy better reflects the actual improvement effect of the maths learner, and we observe that precision accuracy (51.55\%) is 1.55\% higher than global accuracy of MathLearner (50\%), which indicates that our program has some learning ability to deal with the learnt maths problems more effectively.

\section{Discussion}
\subsection{Advantages}
In this thesis, we proposed the "MathLearner" framework, designed to enhance the mathematical problem-solving capabilities of Large Language Models (LLMs). By integrating principles of human learning, specifically inductive reasoning, MathLearner equips LLMs to learn and resolve mathematical challenges effectively.

The framework has two key achievements. First, MathLearner has demonstrated considerable success in boosting LLMs' capabilities in solving mathematical problems, especially those requiring precise reasoning and pattern recognition. By introducing a feature-based retrieval method, the framework effectively enhances the model’s generalization ability, allowing it to approach new and unseen problems with more accurate solutions. The second Strength of the framework is about diverse training data. By leveraging the MATH dataset, MathLearner has access to a broad spectrum of challenging mathematical problems. This diversity is crucial, as it mirrors the real-world complexity and variety of mathematical tasks highlighted in the introduction. The dataset’s comprehensive provision of step-by-step solutions not only deepens the model’s understanding but also enhances its ability to engage in structured mathematical reasoning, a fundamental aspect of developing reliable AI tools in educational and technological applications.

\subsection{Limitations and Future Work}
Although MathLearner has demonstrated an outstanding ability to answer mathematics problems, we note that there are several limitations in our research.
\paragraph{Flaws in simulation of human learning process}
Although MathLearner can store learned solutions in an external database through features, and search through the database to retrieve similar solutions using the encountered problem's feature, the simulation of the human learning process is still not complete. LLMs still do not have the memory of what they have "learned" by storing the solutions in an external database. When encountering a new problem, LLMs still use their pre-trained knowledge to generate features for the problems. Thus, one possible reason why MathLearner can perform better than the baseline is that LLMs have gained impressions of the problems in their pre-trained knowledge. To solve this problem, future work can focus on empowering LLMs to perform real-time updates of knowledge. Using all the possible features to fine-tune LLMs can be a primary solution to this problem.
\paragraph{The ambiguity of the definition of feature}
In this paper, features for one problem contain the category of the problem, like geometry or calculus, they also contain the theorems used in each step of the solution, like the solution formula of quadratic equations with one variable. However, these features are limited compared with features learned by humans. Humans can learn specific structures of problems and use these structural features to put problems into specific question types. All the problems in one question type can share a common solution. It is difficult to generate this type of feature since it is hard to describe this kind of feature through words. Furthermore, the lack of pre-trained knowledge can result in inconsistency of generated features regarding the same structure. Future research can try to design a standardised language for regular words in features and fine-tune LLMs to generate features according to the standardised language of features.
\paragraph{Flaws in the format of modified solution}
In this paper, all the textual solutions will be translated into Python programs. However, for some specific categories of problems, like geometry, it can be difficult to use programs to solve them. In fact, programs are more fittable to be used to do calculations. Thus, a better practice should use both programs and natural language to form modified solutions, and only use programs as tools to do calculation tasks.

\section{Conclusion}
We introduce an agent-based framework designed to tackle mathematical problems through inductive reasoning. This framework mimics the human ability to generalize from learned information and apply it effectively to new problem-solving scenarios. Demonstrated to excel in mathematical reasoning, it has significantly enhanced global accuracy by 20.96\% compared to the baseline Chain of Thought method, and has successfully resolved 17.54\% of the problems that the baseline method failed to solve. Key contributions of MathLearner include its innovative use of feature-based retrieval methods which enhance the model's ability to generalize from learned examples and apply this knowledge effectively to new, similar problems. These advancements have proven particularly effective in educational contexts, where MathLearner can serve as a personalized learning aid, thus democratizing access to quality educational resources.

Despite these advances, the research still points out some limitations. For example, the framework's feature generalisation for mathematical problems was not based on the original learning content, fewer types of features were generalised, and some topics could not be solved using only procedures to represent the solution process. Future developments will focus on extending the adaptive capabilities of LLM by incorporating more dynamic data processing and advanced reasoning techniques. This may involve exploring unsupervised learning paradigms that enable LLMs to independently acquire and apply new knowledge from unstructured data, thus expanding their range of applications.


\printbibliography

\end{document}